\documentclass{article}

\usepackage{PRIMEarxiv}

\usepackage[T1]{fontenc}
\usepackage{graphicx}
\usepackage{orcidlink}
\usepackage{pgfplots}
\usepackage{amsfonts}
\pgfplotsset{compat=1.18}
\usepackage{subcaption}
\usepackage{caption}
\usepackage{booktabs}
\usepackage{threeparttable}
\usepackage{algorithm}
\usepackage{algorithmic}
\usepackage{amsmath}
\usepackage{cleveref}
\DeclareMathOperator*{\argmin}{arg\,min}
\usepackage{stfloats}
\usepackage{pifont}
\newcommand{\cmark}{\ding{51}}

\title{Unsupervised Learning for Missing Modalities in Multimodal Learning
}

\author{
  Hassan Ismkhan* and Hamid Bouchahcia \\
  Bournemouth University \\
  Bournemouth, UK \\
  \{hismkhan*, abouchachia\}@bournemouth.ac.uk \\ \\
  *Responsible author
}

\begin{document}
\begin{large}  \begin{center}In the name of God\end{center} 
\end{large} 

\vspace{2\baselineskip}
\maketitle              
\begin{abstract}
    This paper addresses the missing-modality challenge in multi-modal learning by introducing Unsupervised Learning for Missing Modalities in Multi-Modal Learning (UL4M4), a flexible framework that imputes missing feature embeddings in a task-independent manner before supervised prediction. We propose modality-specific 
    normalization and a novel partial-modality distance metric to enable fair clustering of incomplete observations, capturing cross-modal structures while preserving scale-invariance across varying dimensionalities and modality counts. Cluster centers from this unsupervised stage guide an iterative greedy imputation process for any 
    missing modalities during training or inference, supporting arbitrary numbers of modalities and arbitrary missing patterns per sample. The imputation module is lightweight, uses frozen encoders, and decouples from the downstream task, allowing easy integration with any fusion/prediction architecture. Extensive experiments under diverse and highly incomplete regimes demonstrate UL4M4's robustness, achieving, to 
    the best of our knowledge, the first consistent F1-Micro scores above 0.7 on challenging missing configurations even when more than 50\% of modality slots are missing. Results are also stable across cluster sizes and significantly outperform state-of-the-art baselines. Code is available here: https://github.com/h-ismkhan/Multimodal-Learning-with-Missing-Modalities-via-Unsupervised-Learning.

\end{abstract}
\section{Introduction}

Multimodal learning has emerged as a powerful paradigm for leveraging complementary 
information from multiple data sources such as text, audio, and visual modalities, 
leading to more robust and accurate predictions across various tasks including 
sentiment analysis, emotion recognition, and classification 
\cite{baltrusaitis2018multimodal,zadeh2017tensor,tsai2019multimodal}. In real-world 
applications, however, data completeness is rarely guaranteed: modalities can be 
missing due to sensor failures, privacy restrictions, communication errors, or 
acquisition constraints \cite{ma2021smil,wang2023multi,ma2021are}. This 
missing-modality problem severely degrades performance in traditional multimodal 
models that assume all modalities are present during both training and inference 
\cite{lin2023missmodal,liu2024modality}.

Existing approaches to handle missing modalities fall into two main categories. 
Generative methods attempt to reconstruct or impute missing modalities using available 
ones \cite{cai2019deep,ma2021smil,liu2024modality,dai2024unbiased}. Non-generative 
methods focus on learning shared and modality-specific representations or adapting 
fusion strategies to maintain robustness under incomplete inputs 
\cite{wang2023multi,ma2021are,lee2023multimodal,jin2023rethinking,yu2021learning,
li2024simmlm}. Several works have explored cross-modal interactions, representation 
alignment, and modality translation to mitigate the impact of missing data 
\cite{wu2021text,fu2024hybrid,chen2025temsa,lin2024dynamically,lin2023missmodal,
zhou2021visual,liu2023social,you2016cross,huang2023cross}. More recently, 
mixture-of-experts architectures with dynamic gating have been proposed to 
adaptively weight modality contributions at both training and inference time 
\cite{li2025simmlm}, and retrieval-augmented in-context learning has been explored 
to address the compounded challenge of missing modalities under data scarcity 
\cite{zhi2024borrowing}.

Despite significant progress, most prior methods suffer from one or more of the 
following limitations: (i) they are restricted to a fixed number of modalities, 
(ii) they handle only specific missing-modality patterns (e.g., one or two modalities 
missing), (iii) imputation or adaptation mechanisms are tightly coupled with the 
downstream task, limiting generalizability --- for instance, expert-based approaches 
\cite{li2025simmlm} require modality-specific network branches and two-stage 
co-training that are not straightforwardly transferable across tasks, while 
retrieval-augmented methods \cite{zhi2024borrowing} depend on the availability of 
full-modality reference samples at inference time, a condition that may not hold in 
severely incomplete regimes, or (iv) they are not rigorously evaluated under severe 
missing-modality regimes. These constraints hinder the deployment of multimodal models 
in highly heterogeneous and unreliable real-world environments.

In this paper, we propose \textbf{UL4M4} (Unsupervised Learning for Multi-Modal 
Missing Modalities), a novel framework that addresses these challenges through the 
following key contributions:

\begin{itemize}
    \item \textbf{Arbitrary number of modalities}: UL4M4 supports datasets with any 
    number of input modalities without requiring architectural redesign.
    \item \textbf{Arbitrary missing patterns at sample level}: The method handles any 
    combination of missing modalities for each training, validation, or test sample, 
    including extremely severe cases.
    \item \textbf{Task-independent unsupervised imputation}: Feature embeddings for 
    missing modalities are imputed via an unsupervised learning stage that is decoupled 
    from the final supervised task prediction, enabling the unsupervised component to 
    be reused or extended to other multimodal learning problems with severe missing data.
    \item \textbf{Rigorous evaluation under severe missing regimes}: Experiments are 
    conducted across a wide range of configurations with very high missing rates --- 
    frequently more than 50\% of all modality slots missing across the dataset.
    \item \textbf{State-of-the-art performance in challenging settings}: On particularly 
    difficult multimodal sentiment analysis benchmarks, UL4M4 achieves, to the best of 
    our knowledge, the first reported F1-Micro scores consistently above 0.7 under such 
    severe missing-modality conditions.
\end{itemize}

Extensive experiments across standard multimodal datasets demonstrate the 
effectiveness, generality, and robustness of UL4M4 under diverse and severe 
missing-modality scenarios, outperforming strong baselines and setting a new reference 
for handling real-world incomplete multimodal data.

The remainder of this paper is organized as follows. ~\cref{section:Related-Work} 
reviews the related work. ~\cref{section:proposal} presents our proposed UL4M4 
framework. ~\cref{section:reporting-results} details the experimental setup, results, 
and analysis. Finally, ~\cref{section:Conclusion} concludes the paper.

\section{Related Work}
\label{section:Related-Work}

Multimodal learning leverages complementary information from multiple modalities, 
like text, audio or visual, to achieve more robust representations than unimodal 
methods \cite{baltrusaitis2018multimodal,zadeh2017tensor}. However, real-world 
applications frequently encounter missing modalities due to sensor failures, privacy 
constraints, or data acquisition issues.

Early approaches assumed complete modalities during both training and inference 
\cite{zadeh2017tensor,tsai2019multimodal}. Recent methods address this challenge in 
two main ways: generative and non-generative.

Generative approaches reconstruct missing modalities from available ones. 
\cite{cai2019deep} introduced deep adversarial learning for multi-modality missing 
data completion. \cite{ma2021smil} proposed SMIL, a Bayesian meta-learning method to 
approximate full-modality embeddings under severely missing conditions. 
\cite{liu2024modality} presented a modality translation-based model using Transformers 
to fuse translated features. \cite{dai2024unbiased} developed MD²N, a duplex 
diffusion network for unbiased recovery through cross-modal transfer. 
\cite{zhao2021missing} proposed the Missing Modality Imagination Network to predict 
missing representations.

Non-generative methods focus on learning shared and modality-specific features to 
maintain robustness. \cite{wang2023multi} used auxiliary tasks for distribution 
alignment and domain classification. \cite{ma2021are} evaluated Transformer 
robustness to missing modalities and optimized fusion strategies. 
\cite{lee2023multimodal} adapted pre-trained Transformers via prompt learning. 
\cite{jin2023rethinking} decomposed prototypes with low-rank constraints for decoding 
under arbitrary missing conditions. \cite{yu2021learning} applied self-supervised 
multi-task learning for modality-specific representations. \cite{li2024simmlm} 
introduced dynamic modality experts and ranking losses for improved robustness. A 
related line of work proposes mixture-of-experts architectures with learnable gating 
to dynamically reweight modality contributions: \cite{li2025simmlm} introduced the 
Dynamic Mixture of Modality Experts (DMoME) together with a More-vs-Fewer ranking 
loss that encourages performance to improve monotonically as more modalities become 
available, demonstrating strong results on segmentation and classification benchmarks. 
Complementarily, \cite{zhi2024borrowing} proposed a retrieval-augmented in-context 
learning framework that borrows full-modality context from nearest neighbours to 
support both missing-modality and low-data regimes, showing particular effectiveness 
when labelled training samples are scarce.

In multimodal sentiment analysis (MSA), several works tackle missing modalities 
through cross-modal interactions and representation alignment: \cite{wu2021text} 
distinguished shared and private semantics via cross-modal prediction; 
\cite{fu2024hybrid} proposed hybrid cross-modal interaction learning; 
\cite{chen2025temsa} enhanced non-linguistic modalities with text-guided mapping; 
\cite{lin2024dynamically} dynamically shifted representations with hybrid-modal 
attention; \cite{lin2023missmodal} aligned modal-missing and modal-complete 
representations; \cite{zhou2021visual} employed hierarchical cross-modality 
interactions; \cite{liu2023social} modeled cross-modal consistency with knowledge 
distillation; \cite{you2016cross} used consistent regression for joint visual-textual 
sentiment; \cite{huang2023cross} focused on interactive cross-modality learning.

Some approaches developed in federated contexts also propose multimodal handling under 
missing modalities: \cite{bao2023multimodal} used prototype masks and contrastive 
learning; \cite{le2025cross} introduced cross-prototype learning with regularization; 
\cite{che2024leveraging} leveraged pre-trained models for modality completion.

While prior works primarily target supervised settings, our UL4M4 advances multimodal 
learning under missing modalities with a unified representation strategy, evaluated on 
standard MSA benchmarks.
\section{Cluster-Guided Iterative Imputation for Missing Modalities}
    \label{section:proposal}
    Let $\mathcal{D} = \{(x_i^{(1)}, \dots, x_i^{(K)}, y_i)\}_{i=1}^N$ denote the multi-modal training dataset, where $x_i^{(m)} \in \mathcal{X}_m$ is the raw input of modality $m$ (or missing) and $y_i$ is the target label. For each modality $m$, a frozen pretrained encoder $E_m : \mathcal{X}_m \to \mathbb{R}^{d_m}$ produces the embedding $\mathbf{z}_i^{(m)} = E_m(x_i^{(m)})$ whenever $x_i^{(m)}$ is available.
    
    To handle missing modalities in this setting, we propose UL4M4, a two-stage unsupervised approach that relies on clustering followed by iterative greedy completion. The method operates entirely on the frozen encoder representations without updating any encoder parameters. In the first stage, we perform multi-modal clustering to obtain cluster centres that capture shared structures across partial observations. In the second stage, we impute missing embeddings for each incomplete sample using an iterative selection of candidate completions derived from these centres. Once all training samples are completed, downstream task training proceeds by fusing the full modality representations and minimizing the task loss using only fusion and head parameters. At inference, test samples with missing modalities are similarly imputed using the training-derived centres and statistics before fusion and prediction.
    
    \subsection{Stage 1: Multi-Modal Clustering with Partial-Modality Distance}
    
    Prior to clustering, we compute modality-specific normalization statistics. For each modality $m$, let $\mathcal{I}_m = \{i : x_i^{(m)} \text{ is available}\}$ be the set of samples containing modality $m$. We calculate the mean and standard deviation over all available embeddings of that modality:
    \begin{equation}
    \boldsymbol{\mu}_m = \frac{1}{|\mathcal{I}_m|} \sum_{i \in \mathcal{I}_m} \mathbf{z}_i^{(m)}, \qquad
    \boldsymbol{\sigma}_m = \sqrt{\frac{1}{|\mathcal{I}_m|} \sum_{i \in \mathcal{I}_m} \|\mathbf{z}_i^{(m)} - \boldsymbol{\mu}_m\|^2}.
    \label{eq:norm-stats}
    \end{equation}
    Normalized features are then obtained via $\tilde{\mathbf{z}}_i^{(m)} = (\mathbf{z}_i^{(m)} - \boldsymbol{\mu}_m) / \boldsymbol{\sigma}_m$ (element-wise) for every available $\mathbf{z}_i^{(m)}$.
    
    We perform $k$-means clustering on the training set using these normalized features. The algorithm is initialized by randomly selecting $k$ training samples as initial cluster centres $\mathbf{c}_1^{(0)}, \dots, \mathbf{c}_k^{(0)}$. Each centre inherits exactly the modalities present in its initializing sample. The clustering proceeds iteratively with the following steps until convergence or a maximum number of iterations is reached.
    
    In the assignment step, every training sample $i$ is assigned to the cluster $j$ that minimizes the partial-modality distance to centre $\mathbf{c}_j$. Let $\mathcal{M}_{i,j}$ be the set of modalities available in both sample $i$ and centre $\mathbf{c}_j$. The distance is defined as
    \begin{equation}
    d(i,j) = \sqrt{ \frac{1}{|\mathcal{M}_{i,j}|} \sum_{m \in \mathcal{M}_{i,j}} \frac{\|\tilde{\mathbf{z}}_i^{(m)} - \tilde{\mathbf{c}}_j^{(m)}\|_2^2 }{d_m} },
    \label{eq:partial-distance}
    \end{equation}
    where $d_m = \dim(\mathbf{z}_i^{(m)}) = \dim(\tilde{\mathbf{c}}_j^{(m)})$ is the dimensionality of modality $m$ and $\tilde{\mathbf{c}}_j^{(m)}$ denotes the normalized embedding of modality $m$ in centre $j$ (if available). If $\mathcal{M}_{i,j} = \emptyset$, the pair is considered infinitely distant.
    
    The two normalization factors in \cref{eq:partial-distance} serve distinct but complementary purposes. First, the outer division by $|\mathcal{M}_{i,j}|$ ensures that distances are comparable across pairs that share different numbers of modalities; without it, pairs sharing more modalities would tend to produce systematically larger summed squared errors (even when per-modality disagreement is small), biasing assignments toward centres with fewer overlaps. Second, the inner division by $d_m$ converts each modality's squared error into a mean squared error per dimension. This is necessary because different modalities typically have very different embedding dimensionalities (e.g., $d_m = 768$ for text versus $d_m = 2048$ or higher for video or audio features). Without this term, high-dimensional modalities would disproportionately influence the overall distance simply by contributing more summation terms, even after per-modality z-score normalization. Together, these factors yield a balanced, scale-invariant, and modality-fair distance that better reflects true semantic disagreement rather than artifacts of overlap cardinality or embedding size.
    
    In the update step, for each cluster $j$ and each modality $m$, the new centre embedding $\mathbf{c}_j^{(m)}$ is computed as the mean of the features over all samples currently assigned to cluster $j$ that possess modality $m$:
    \begin{equation}
    \mathbf{c}_j^{(m)} = \frac{1}{|\mathcal{I}_j^{(m)}|} \sum_{i \in \mathcal{I}_j^{(m)}} \tilde{\mathbf{z}}_i^{(m)},
    \label{eq:centre-update}
    \end{equation}
    where $\mathcal{I}_j^{(m)}$ is the set of assigned samples that contain modality $m$. If no assigned sample possesses modality $m$, then $\mathbf{c}_j^{(m)}$ remains undefined (missing) in the centre.
    
    After convergence, we obtain $k$ cluster centres $\{\mathbf{c}_j\}_{j=1}^k$, each potentially missing some modalities, together with a hard assignment of every training sample to one of these clusters. The procedure is summarized in \cref{alg:clustering}.
    
    \begin{algorithm}[tb]
    \caption{UL4M4 Clustering Stage (Stage 1)}
    \label{alg:clustering}
    \begin{algorithmic}[1]
    \REQUIRE Dataset $\mathcal{D}$, number of clusters $k$, max iterations $T$
    \STATE Compute normalization statistics $\{\boldsymbol{\mu}_m, \boldsymbol{\sigma}_m\}$ for each modality $m$ using \cref{eq:norm-stats}
    \STATE Normalize all available embeddings to obtain $\{\tilde{\mathbf{z}}_i^{(m)}\}$
    \STATE Randomly select $k$ samples as initial centres $\{\mathbf{c}_j^{(0)}\}_{j=1}^k$
    \FOR{$t = 1$ to $T$}
        \STATE \emph{Assignment Step:}
        \FOR{each sample $i$}
            \STATE Assign $i$ to cluster $j^* = \argmin_j d(i,j)$ using \cref{eq:partial-distance}
        \ENDFOR
        \STATE \emph{Update Step:}
        \FOR{each cluster $j$ and modality $m$}
            \IF{$\mathcal{I}_j^{(m)} \neq \emptyset$}
                \STATE Update $\mathbf{c}_j^{(m)}$ using \cref{eq:centre-update}
            \ELSE
                \STATE Leave $\mathbf{c}_j^{(m)}$ undefined
            \ENDIF
        \ENDFOR
        \IF{assignments converged}
            \STATE Break
        \ENDIF
    \ENDFOR
    \RETURN Cluster centres $\{\mathbf{c}_j\}_{j=1}^k$ and assignments
    \end{algorithmic}
    \end{algorithm}
    
    \subsection{Stage 2: Iterative Greedy Completion of Training Samples}
    
    For each training sample $i$ that is missing at least one modality, we first construct an initial set of candidate completions $\mathcal{C}_i = \{\mathbf{d}_{i,1}, \dots, \mathbf{d}_{i,k}\}$. For every cluster centre index $f \in \{1,\dots,k\}$, the candidate $\mathbf{d}_{i,f}$ is formed by taking all modalities originally available in sample $i$ and, for each missing modality $m$, filling it with the corresponding embedding $\mathbf{c}_f^{(m)}$ from centre $f$ whenever $\mathbf{c}_f^{(m)}$ is defined. If centre $f$ does not possess a required missing modality $m$, then modality $m$ remains absent in $\mathbf{d}_{i,f}$.
    
    The completion process then proceeds iteratively. In each iteration, we compute the partial-modality distance (as defined in \cref{eq:partial-distance}) between every remaining candidate $\mathbf{d} \in \mathcal{C}_i$ and every cluster centre $\mathbf{c}_j$ ($j=1,\dots,k$), using the z-score normalized versions of the embeddings present in both the candidate and the centre. If no modalities are shared between a candidate and a centre ($\mathcal{M}_{\mathbf{d},j} = \emptyset$), the distance is considered infinite.
    
    We identify the pair $(\mathbf{d}^*, j^*)$ that achieves the global minimum distance among all candidate-centre pairs. The selected candidate $\mathbf{d}^*$ is removed from $\mathcal{C}_i$. For every modality $m$ that is still missing in the target representation of sample $i$, we copy the embedding $\mathbf{d}^{*(m)}$ into the representation of $i$ (if modality $m$ is present in $\mathbf{d}^*$).
    
    This greedy selection-and-imputation step is repeated until the representation of sample $i$ contains embeddings for all $K$ modalities. Because the union of all cluster centres collectively covers every modality (assuming the training data contains all modalities overall) and because every centre is used to generate at least one initial candidate, the process is guaranteed to terminate with a fully completed representation. The procedure is summarized in \cref{alg:completion}.
    
    \begin{algorithm}[tb]
    \caption{UL4M4 Completion Stage (Stage 2) for a Sample $i$}
    \label{alg:completion}
    \begin{algorithmic}[1]
    \REQUIRE Cluster centres $\{\mathbf{c}_j\}_{j=1}^k$, incomplete sample $i$ with available embeddings $\{\mathbf{z}_i^{(m)}\}$
    \STATE Initialize $\mathcal{C}_i = \emptyset$
    \FOR{$f = 1$ to $k$}
        \STATE Create candidate $\mathbf{d}_{i,f}$: copy available modalities from $i$; fill missing $m$ with $\mathbf{c}_f^{(m)}$ if defined
        \STATE Add $\mathbf{d}_{i,f}$ to $\mathcal{C}_i$
    \ENDFOR
    \WHILE{sample $i$ has missing modalities}
        \STATE Compute distances $d(\mathbf{d}, j)$ for all $\mathbf{d} \in \mathcal{C}_i$, $j=1,\dots,k$ using \cref{eq:partial-distance}
        \STATE Find $(\mathbf{d}^*, j^*) = \argmin d(\mathbf{d}, j)$
        \STATE Remove $\mathbf{d}^*$ from $\mathcal{C}_i$
        \FOR{each missing modality $m$ in $i$}
            \IF{$m$ is present in $\mathbf{d}^*$}
                \STATE Copy $\mathbf{d}^{*(m)}$ to $i$
            \ENDIF
        \ENDFOR
    \ENDWHILE
    \RETURN Completed representation of sample $i$
    \end{algorithmic}
    \end{algorithm}
    
    \subsection{Downstream Task Training and Inference}
    
    Once every training sample has a complete set of $K$ modality embeddings (either originally available or imputed via UL4M4), a fusion module combines the $K$ (now complete) modality representations into a single cross-modal representation, which is then passed to the task-specific head to produce the final prediction. 
    The fusion module $f_\mathcal{S}: \prod_{i \in \mathcal{S}} \mathbb{R}^{d_i} \to \mathbb{R}^{d_f}$ uses multi-head self-attention (MHSA) for cross-modal interactions. Inputs are projected onto $d_f$:
    \begin{equation} \label{eq:2}
    \mathbf{m}_i^{\text{proj}} = W_i^{\text{proj}} \mathbf{M}_i' + \mathbf{b}_i^{\text{proj}}, \quad i \in \mathcal{S}.
    \end{equation}
    Stacking into sequence $\mathbf{M} = [\mathbf{m}_1^{\text{proj}}; \dots; \mathbf{m}_{|\mathcal{S}|}^{\text{proj}}] \in \mathbb{R}^{|\mathcal{S}| \times d_f}$ and applying $L$ MHSA layers yields:
    \begin{equation} \label{eq:3}
    \text{head}_h = \text{softmax}\left( \frac{(\mathbf{M} W_h^Q) (\mathbf{M} W_h^K)^\top}{\sqrt{d_f / H}} \right) (\mathbf{M} W_h^V),
    \end{equation}
    \begin{equation} \label{eq:4}
    \text{MHSA}(\mathbf{M}) = \text{Concat}(\text{head}_1, \dots, \text{head}_H) W^O,
    \end{equation}
    where attention weights in each head are additionally regularized via dropout with probability $p$. Following the standard Transformer block convention~\cite{vaswani2017attention}, dropout is applied to the attention output prior to the residual connection and layer normalization:
    \begin{equation} \label{eq:5}
    \mathbf{M}^{(\ell)} = \text{LayerNorm}\!\left(\mathbf{M}^{(\ell-1)} + \text{Dropout}_p\!\left(\text{MHSA}(\mathbf{M}^{(\ell-1)})\right)\right).
    \end{equation}
    Finally, mean pooling is applied over the modality dimension, $\mathbf{F}_\mathcal{S} = \frac{1}{|\mathcal{S}|} \sum_i \mathbf{M}_i^{(L)}$, followed by an output projection and activation:
    \begin{equation} \label{eq:6}
    f_\mathcal{S}(\{\mathbf{M}_i'\}_{i \in \mathcal{S}}) = \text{ReLU}(W^{\text{out}} \mathbf{F}_\mathcal{S} + \mathbf{b}^{\text{out}}).
    \end{equation}
    
    This attention-based fusion with dropout dynamically weights modalities while reducing overfitting and enhancing robustness to noisy or incomplete inputs.
    During training we minimize the task loss $\mathcal{L}_{\text{task}}$ (cross-entropy for classification or mean squared error for regression) over the entire dataset using only the task-related parameters (fusion and head), while all encoders and the imputed embeddings remain fixed. At inference time, missing modalities in a test sample are imputed using the same two-stage UL4M4 procedure (with cluster centres and normalization statistics computed solely from the training set), after which the complete representation is fused and passed through the trained task head.

\section{Results of Experiments}
\label{section:reporting-results}

In experiments we use \textbf{MM-IMDb} dataset which comprises image (movie posters) and text (plot summaries) modalities for movie genre classification across 27 distinct genres including Drama, Comedy, Thriller, and Romance. Each sample can be associated with multiple genres simultaneously, making this a challenging multi-label classification task. For missing modalities, we various availability patterns:
\begin{itemize}
    \item \textit{$\alpha$-image-$\beta$-text}: $\alpha$\% image with $\beta$\% text,
    \item \textit{complex-$\gamma$-$\alpha$-$\beta$}: $\gamma$\% both-modalities-presence, $\alpha$\% image only and $\beta$\% text only. 
\end{itemize}
Other dataset used in experiments is \textbf{CMU-MOSI} which is a multi-modal Opinion Sentiment Intensity dataset and contains three modalities: text (transcripts), audio (waveforms), and video (facial expressions) from YouTube movie review segments. Each segment is annotated with a continuous sentiment score ranging from -3 (strongly negative) to +3 (strongly positive), forming a regression task. Missing modality configurations specify availability percentages for each modality independently or complex distributions defining proportions for all-three-present, single-modality, and pairwise-modality combinations:
\begin{itemize}
    \item \textit{X-text-Y-audio-Z-video}: \textit{X}\% text, \textit{Y}\% audio, and \textit{Z}\% video,
    \item \textit{complex-TAV-T-A-V-TA-TV-AV}: \textit{TAV}\% all-modalities-presence, \textit{T}\% text only, \textit{A}\% audio only, \textit{V}\% video only, \textit{TA}\% text with audio but no video, \textit{TV}\% text with video but no audio, \textit{AV}\% audio with video but no text.    
\end{itemize}

The Sleep-EDF polysomnography is another dataset used in experiments,  which has $K=5$ signal modalities, 5 sleep-stage classes (W, N1, N2, N3, REM), and signals are resampled to 250 time steps. Two parameters $ps$ and $pm$ are used for missing modalities configuration. The $ps$ shows how fraction(\%) of samples face with missing modalities, the $pm$ shows what fraction of modalities are missing in these samples.

    CMU-MOSI remains a challenging benchmark, especially under missing-modality settings. While general missing-modality methods such as M$^3$Care\cite{zhang2022m3care}, ShaSpec\cite{wang2023multi}, and PmcmFL\cite{bao2023multimodal} excel in domains like healthcare, general computer vision datasets, or federated learning, they are rarely evaluated on CMU-MOSI. Therefore, we focus comparisons on MissModal, as it is specifically designed for CMU-MOSI and outperforms other state-of-the-art methods.
    In these experiments M$^3$Care\cite{zhang2022m3care}, ShaSpec\cite{wang2023multi}, ICL-CA\cite{zhi2024borrowing}, and SimMLM\cite{li2024simmlm} are participated. To show UL4M4 is also comparable against most task specific SOTA, for CMU-MOSI, UL4M4 also competes with MissModal\cite{lin2023missmodal}, which is specifically designed for sentimental analysis and specially this dataset.
    For all experiments in this section, 10 independent runs are performed. 

\begin{table*}[t]
\centering
\caption{F-micro Obtained by the Modles on MM-IMDb and CMU-MOSI}
\vspace{-0.0em}
\label{tab:fmicro}
\resizebox{\textwidth}{!}{%
\begin{tabular}{l|cccccc|cccccccc}
\toprule
& \multicolumn{6}{c|}{\textbf{MM-IMDb}} & \multicolumn{8}{c}{\textbf{CMU-MOSI}} \\
\cmidrule(lr){2-7} \cmidrule(lr){8-15}
Method
& \rotatebox{90}{\scriptsize 100-image-100-text}
& \rotatebox{90}{\scriptsize 100-image-20-text}
& \rotatebox{90}{\scriptsize 20-image-100-text}
& \rotatebox{90}{\scriptsize complex-10-45-45}
& \rotatebox{90}{\scriptsize complex-20-40-40}
& \rotatebox{90}{\scriptsize complex-30-35-35}
& \rotatebox{90}{\scriptsize 100-text-100-audio-100-video}
& \rotatebox{90}{\scriptsize 100-text-100-audio-20-video}
& \rotatebox{90}{\scriptsize 100-text-20-audio-100-video}
& \rotatebox{90}{\scriptsize 100-text-20-audio-20-video}
& \rotatebox{90}{\scriptsize 20-text-100-audio-100-video}
& \rotatebox{90}{\scriptsize 20-text-100-audio-20-video}
& \rotatebox{90}{\scriptsize 20-text-20-audio-100-video}
& \rotatebox{90}{\scriptsize complex-20-20-20-10-10-10-10} \\
\midrule
ShaSpec       & 0.49 & 0.39 & 0.45 & 0.39 & 0.38 & 0.43 & 0.35 & 0.33 & 0.32 & 0.30 & 0.28 & 0.25 & 0.27 & 0.29 \\
$M^3$Care     & 0.46 & 0.39 & 0.43 & 0.43 & 0.37 & 0.45 & 0.36 & 0.34 & 0.33 & 0.31 & 0.29 & 0.26 & 0.28 & 0.30 \\
ICL-CA      & 0.56 & 0.48 & 0.43 & 0.46 & 0.47 & 0.43   & 0.71 & 0.56 & 0.78 & 0.56 & 0.67 & 0.56 & 0.65 & 0.68 \\
SimMLM         & 0.58 & 0.54 & 0.54 & 0.51 & 0.51 & 0.54   & 0.77 & 0.76 & 0.77 & 0.45 & 0.69 & 0.54 & 0.63 & 0.68 \\
UL4M4          & \textbf{0.93} & \textbf{0.92} & \textbf{0.92} & \textbf{0.92} & \textbf{0.92} & \textbf{0.92} & \textbf{0.81} & \textbf{0.81} & \textbf{0.81} & \textbf{0.81} & \textbf{0.69} & \textbf{0.59} & \textbf{0.69} & \textbf{0.72} \\
\midrule
Winner (t-test)        & \cmark  & \cmark & \cmark & \cmark & \cmark & \cmark & \cmark & \cmark & \cmark & \cmark & \cmark & \cmark & \cmark & \cmark \\
Winner (WCOX)        & \cmark & \cmark & \cmark & \cmark & \cmark & \cmark & \cmark & \cmark & \cmark & \cmark & \cmark & \cmark & \cmark & \cmark \\
\bottomrule
\end{tabular}%
}
\vspace{-1.0em}
\end{table*}

\subsection{The Basic Results}

    Tables~\ref{tab:fmicro} and~\ref{tab:edf-comparison} summarize the F-Micro and accuracy results across MM-IMDb, CMU-MOSI, and Sleep-EDF, respectively.
    
    \paragraph{MM-IMDb.}
    UL4M4 achieves substantially higher F-Micro scores than all competing methods across every missing-modality configuration on MM-IMDb, and both t-test and Wilcoxon signed-rank test confirm the differences are statistically significant in all cases. The margin over the second-best baseline is consistently large, suggesting that the cluster-guided imputation strategy is particularly well-suited to this bimodal setting, where text and image embeddings are highly complementary and the downstream task is multi-label classification.
    
    \paragraph{CMU-MOSI.}
    On CMU-MOSI, UL4M4 again outperforms all baselines across all configurations in F-Micro, with statistical significance confirmed by both tests. The advantage is most pronounced in the configurations where text availability is severely reduced, which are the hardest settings for sentiment analysis given the central role of language. In the less severe configurations — where all three modalities are at least partially available — UL4M4 maintains its edge while delivering consistently strong performance across the full range of missing-modality patterns.
    
\begin{table*}[t]
\centering
\caption{Comparison of Accuracy of Models on 5-Modal EDF Dataset}
\label{tab:edf-comparison}
\small
\resizebox{\textwidth}{!}{%
\begin{tabular}{l cccc cccc}
\toprule
& \multicolumn{4}{c}{\textbf{$pm=0.4$}}
& \multicolumn{4}{c}{\textbf{$pm=0.6$}} \\

\cmidrule(lr){2-5}
\cmidrule(lr){6-9}
\textbf{Missing Ratio}
& \textbf{ps=0.2}
& \textbf{ps=0.4}
& \textbf{ps=0.6}
& \textbf{ps=0.8}
& \textbf{ps=0.2}
& \textbf{ps=0.4}
& \textbf{ps=0.6}
& \textbf{ps=0.8} \\
\midrule
ShaSpec& 0.32& 0.34& 0.34& 0.34& 0.33& 0.33& 0.32& 0.32\\
$M^3$Care & 0.39& 0.39& 0.38& 0.37& 0.38& 0.37& 0.36&0.36\\
ICL-CA & 0.39 & 0.36 & 0.36 & 0.35& 0.36 & 0.36 & 0.35 &0.35 \\
SimMLM & 0.39 & 0.38 & 0.38 & 0.38& 0.37 & 0.36 & 0.35 &0.34 \\
UL4M4 & 0.43 & 0.43 & 0.41 & 0.42& 0.42 & 0.41 & 0.41 & 0.40 \\
\midrule
Winner(t-test)
& \cmark & \cmark & \cmark & \cmark
& \cmark & \cmark & \cmark & \cmark \\
Winner(WCOX)
& \cmark & \cmark & \cmark & \cmark
& \cmark & \cmark & \cmark & \cmark \\
\bottomrule
\end{tabular}%
}
\end{table*}
    
    \paragraph{Sleep-EDF.}
    On the five-modal Sleep-EDF dataset, UL4M4 consistently outperforms all baselines across the full range of missing-sample and missing-modality-fraction combinations considered, again with statistical significance confirmed by both tests. Importantly, performance degrades gracefully as missing rates increase in both dimensions, whereas the baselines exhibit more pronounced drops under the more severe configurations. This robustness in the high-missing regime underlines the scalability of UL4M4 to datasets with a larger number of modalities and more aggressive missing patterns.

\subsection{Comparison with a Task-Specific State-of-the-Art: UL4M4 vs.\ MissModal}
    
    To further situate UL4M4 within the landscape of existing methods, we compare it against MissModal~\cite{lin2023missmodal}, a method specifically designed for multimodal sentiment analysis and particularly tailored to the CMU-MOSI benchmark. Unlike the general-purpose baselines evaluated above, MissModal incorporates task-specific inductive biases and alignment objectives optimized for this setting, making it a particularly demanding point of comparison for a general framework such as UL4M4. Table~\ref{tab:missmodal-vs-ul4m4} reports MSE, F-Micro, and F-Macro side by side across all evaluated configurations, with bold entries indicating a statistically significant advantage.
    \begin{table}[t]
\centering
\caption{Comparison of MissModal vs.\ UL4M4 across MSE, F-Micro, and F-Macro on CMU-MOSI.}
\label{tab:missmodal-vs-ul4m4}
\footnotesize
\begin{tabular}{lcccccc}
\toprule
 & \multicolumn{2}{c}{MSE $\downarrow$} & \multicolumn{2}{c}{F-Micro $\uparrow$} & \multicolumn{2}{c}{F-Macro $\uparrow$} \\
\cmidrule(lr){2-3}\cmidrule(lr){4-5}\cmidrule(lr){6-7}
Configuration & MissModal & UL4M4 & MissModal & UL4M4 & MissModal & UL4M4 \\
\midrule
100-text-100-audio-100-video & 1.11 & \textbf{1.01} & 0.80 & \textbf{0.81} & 0.80 & \textbf{0.81} \\
100-text-100-audio-20-video  & 1.08 & \textbf{1.03} & 0.80 & \textbf{0.81} & 0.80 & 0.81 \\
100-text-20-audio-100-video  & 1.13 & \textbf{0.99} & 0.80 & \textbf{0.81} & 0.80 & 0.81 \\
100-text-20-audio-20-video   & 1.07 & 1.04          & 0.81 & 0.81          & 0.81 & 0.81 \\
20-text-100-audio-100-video  & 2.04 & \textbf{1.91} & 0.67 & \textbf{0.69} & 0.67 & \textbf{0.68} \\
20-text-100-audio-20-video   & \textbf{2.14} & 2.19 & 0.57 & 0.59          & 0.52 & 0.52 \\
20-text-20-audio-100-video   & 2.00 & \textbf{1.90} & 0.68 & \textbf{0.69} & 0.67 & 0.68 \\
complex-20-20-20-10-10-10-10 & 1.59 & 1.56          & 0.71 & 0.72          & 0.71 & 0.71 \\
\bottomrule
\end{tabular}
\begin{tablenotes}
\small
\item Bold values indicate the statistically significantly better result. Non-bold entries indicate no significant difference.
\end{tablenotes}
\end{table}
    Across the MSE metric, UL4M4 yields significantly lower error than MissModal in the majority of configurations, with MissModal holding a statistically significant advantage only in the most audio-degraded setting. This isolated reversal is consistent with the intuition that severe degradation of a single modality can expose edge cases for any imputation strategy, yet it remains the exception rather than the rule. For F-Micro and F-Macro, UL4M4 matches or significantly outperforms MissModal in most configurations, with no configuration showing a statistically significant advantage for MissModal on either metric.
    
    Taken together, these results demonstrate that UL4M4, despite being a domain-agnostic framework with no task-specific design choices, is competitive with — and in most settings superior to — a method purpose-built for multimodal sentiment analysis on CMU-MOSI. This finding strengthens the case for UL4M4 as a general and robust solution to the missing-modality problem.
    
\begin{figure*}[t]
\centering
\begin{subfigure}[b]{0.32\columnwidth}
    \centering
    \includegraphics[width=\linewidth]{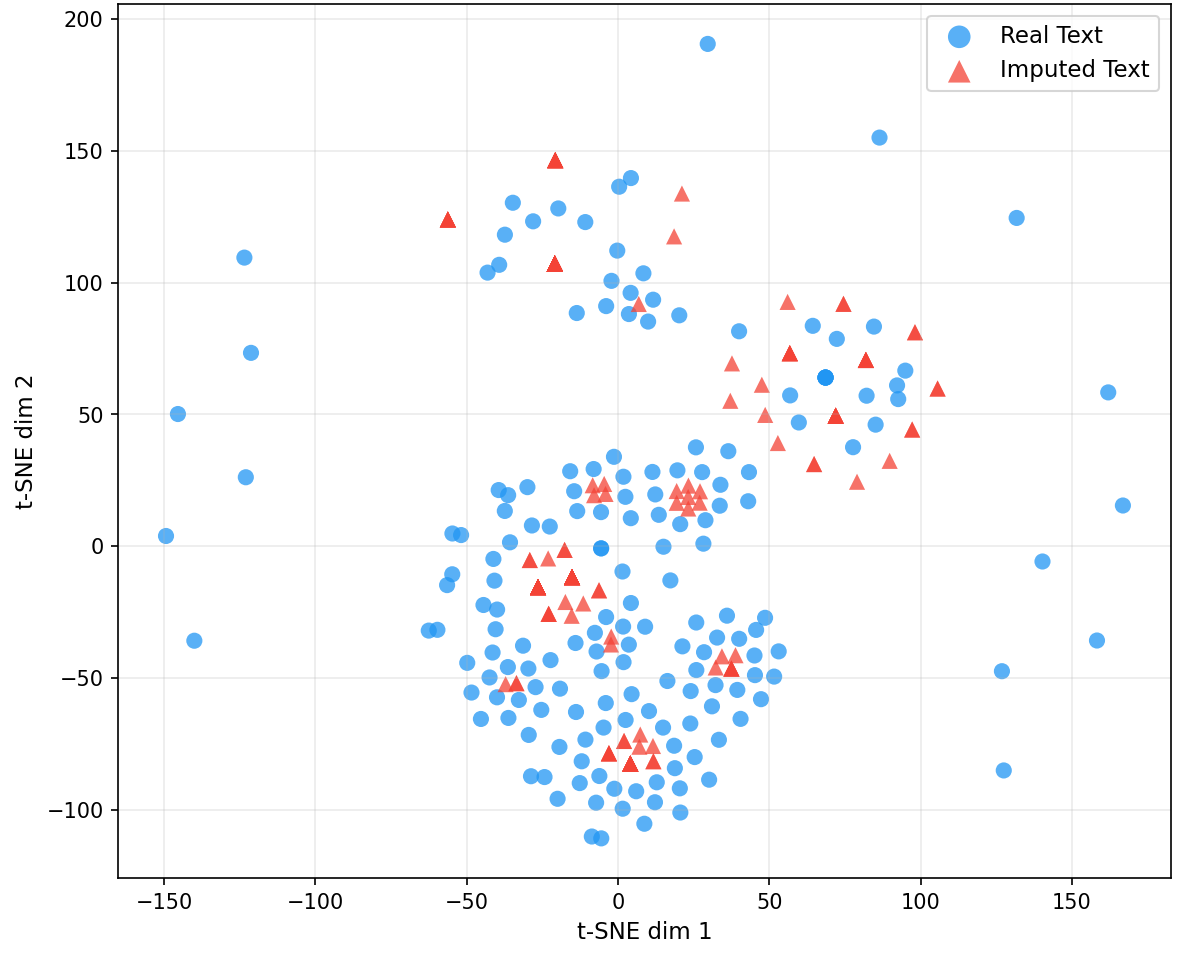}
    \vspace{-0.1em}
    \caption*{Text}
\end{subfigure}
\begin{subfigure}[b]{0.32\columnwidth}
    \centering
    \includegraphics[width=\linewidth]{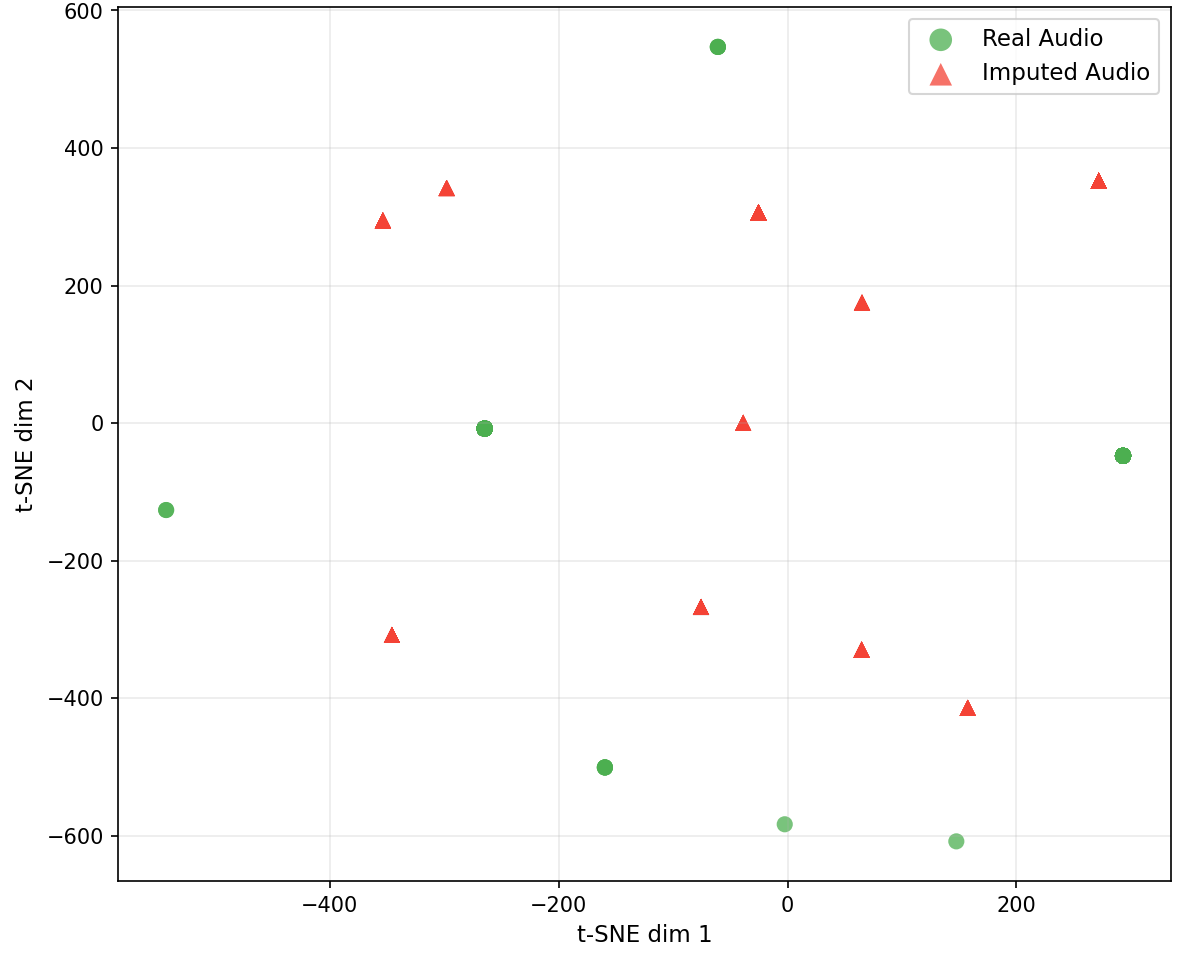}
    \vspace{-0.1em}
    \caption*{Audio}
\end{subfigure}
\begin{subfigure}[b]{0.32\columnwidth}
    \centering
    \includegraphics[width=\linewidth]{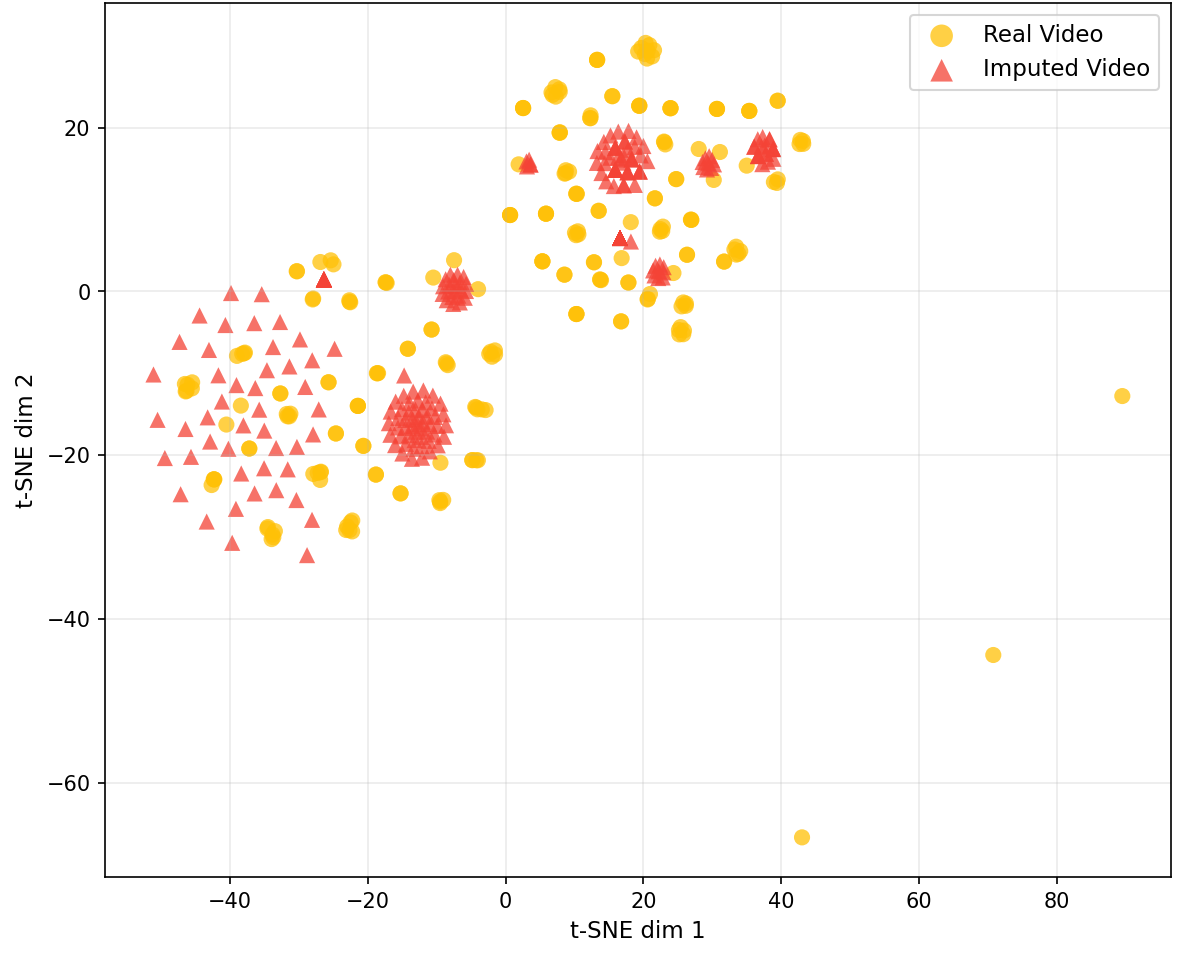}
    \vspace{-0.1em}
    \caption*{Video}
\end{subfigure}

\vspace{-0.3em}

\caption{Visualisation of the feature embedding of the missing modalities for complex-20-20-20-10-10-10-10.}
\vspace{-2.0 em}
\label{fig:tsne-visualization}
\end{figure*}

\cref{fig:tsne-visualization} qualitatively evaluates the quality of imputed embeddings for UL4M4 in the third run under the complex missing-modality setting, \textit{complex-20-20-20-10-10-10-10}, via t-SNE\cite{van2008visualizing} projection to 2D. In each diagram, blue points represent the ground-truth embeddings of text features, green points represent those of audio features, and yellow points represent those of video features; red points denote their corresponding imputed feature embeddings. UL4M4 produces imputed points that tightly overlap and integrate with the ground-truth embeddings across all modalities, preserving realistic distributions. It should be noted that these results are from the test split of the data and that the number of blue, green, and yellow points is equal. However, for the audio modality, the frozen encoder produces nearly identical feature embeddings for several samples; consequently, their imputed embeddings are also similar, causing the audio diagram to appear to contain fewer points than the other two.

\subsection{Impact of Cluster Number on UL4M4 Performance}
To assess the sensitivity of UL4M4 to the number of clusters $  k  $, we vary $  k  $ from 10 to 100 in steps of 10 and evaluate F-Micro scores across various missing modality configurations, with results averaged over 10 independent runs. \cref{fig:effects-of-number-of-clusters} illustrates these results, showing relatively stable performance with minor fluctuations (typically within 0.01-0.02) for each configuration. This indicates that UL4M4 is robust to the choice of $  k  $ in this range, as increasing $  k  $ does not lead to significant improvements or degradations, allowing flexible selection based on computational constraints without compromising effectiveness.

\subsection{Sensitivity to the Number of Clusters $k$}

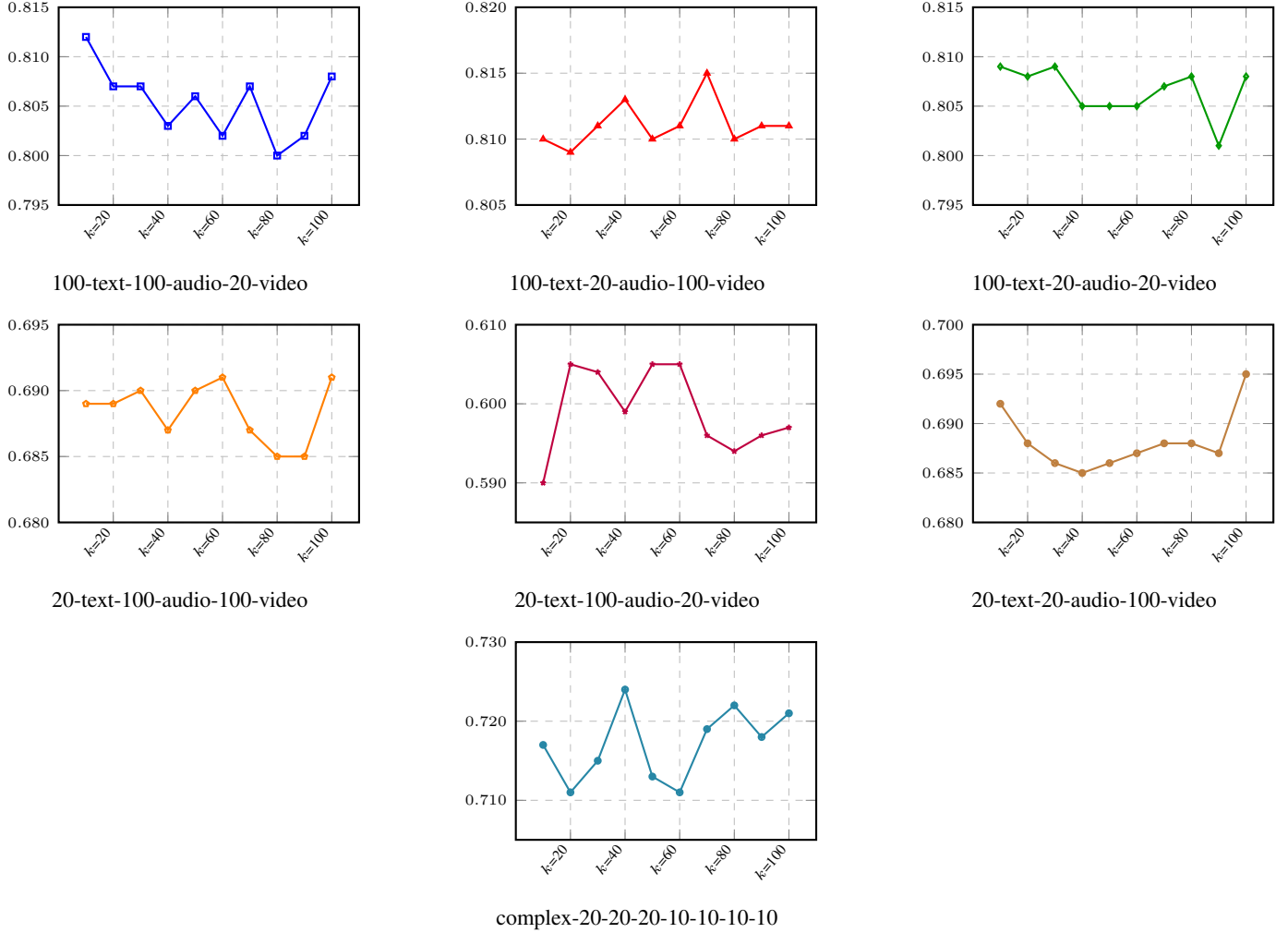
\begin{figure*}[t]
\centering
\makebox[\textwidth][c]{%
\begin{minipage}{1.15\textwidth}
\centering
\begin{subfigure}[b]{0.31\textwidth}
    \centering
    \begin{tikzpicture}
    \begin{axis}[
        xmin=0, xmax=110,
        ymin=0.795, ymax=0.815,
        yticklabel style={/pgf/number format/fixed, /pgf/number format/fixed zerofill, /pgf/number format/precision=3, font=\scriptsize},
        yticklabel style={font=\tiny},
        xtick={20, 40, 60, 80, 100},
        xticklabels={$k$=20,$k$=40,$k$=60,$k$=80,$k$=100},
        xticklabel style={rotate=50, anchor=east, font=\tiny},
        width=\linewidth,
        height=0.75\linewidth,
        grid style=dashed,
        ymajorgrids=true,
        xmajorgrids=true,
        mark size=1.2pt,
        line width=0.8pt,
        tick label style={font=\small},
        label style={font=\small}
    ]
    \addplot[color=blue, mark=square] coordinates {
        (10,0.812)(20,0.807)(30,0.807)(40,0.803)(50,0.806)
        (60,0.802)(70,0.807)(80,0.8)(90,0.802)(100,0.808)
    };
    \end{axis}
    \end{tikzpicture}
    \caption*{100-text-100-audio-20-video}
    \label{fig:config-a}
\end{subfigure}
\hfill
\begin{subfigure}[b]{0.31\textwidth}
    \centering
    \begin{tikzpicture}
    \begin{axis}[
        xmin=0, xmax=110,
        ymin=0.805, ymax=0.82,
        yticklabel style={/pgf/number format/fixed, /pgf/number format/fixed zerofill, /pgf/number format/precision=3, font=\scriptsize},
        yticklabel style={font=\tiny},
        xtick={20, 40, 60, 80, 100},
        xticklabels={$k$=20,$k$=40,$k$=60,$k$=80,$k$=100},
        xticklabel style={rotate=50, anchor=east, font=\tiny},
        width=\linewidth,
        height=0.75\linewidth,
        grid style=dashed,
        ymajorgrids=true,
        xmajorgrids=true,
        mark size=1.2pt,
        line width=0.8pt,
        tick label style={font=\small},
        label style={font=\small}
    ]
    \addplot[color=red, mark=triangle] coordinates {
        (10,0.81)(20,0.809)(30,0.811)(40,0.813)(50,0.81)
        (60,0.811)(70,0.815)(80,0.81)(90,0.811)(100,0.811)
    };
    \end{axis}
    \end{tikzpicture}
    \caption*{100-text-20-audio-100-video}
    \label{fig:config-b}
\end{subfigure}
\hfill
\begin{subfigure}[b]{0.31\textwidth}
    \centering
    \begin{tikzpicture}
    \begin{axis}[
        xmin=0, xmax=110,
        ymin=0.795, ymax=0.815,
        yticklabel style={/pgf/number format/fixed, /pgf/number format/fixed zerofill, /pgf/number format/precision=3, font=\scriptsize},
        yticklabel style={font=\tiny},
        xtick={20, 40, 60, 80, 100},
        xticklabels={$k$=20,$k$=40,$k$=60,$k$=80,$k$=100},
        xticklabel style={rotate=50, anchor=east, font=\tiny},
        width=\linewidth,
        height=0.75\linewidth,
        grid style=dashed,
        ymajorgrids=true,
        xmajorgrids=true,
        mark size=1.2pt,
        line width=0.8pt,
        tick label style={font=\small},
        label style={font=\small}
    ]
    \addplot[color=green!60!black, mark=diamond] coordinates {
        (10,0.809)(20,0.808)(30,0.809)(40,0.805)(50,0.805)
        (60,0.805)(70,0.807)(80,0.808)(90,0.801)(100,0.808)
    };
    \end{axis}
    \end{tikzpicture}
    \caption*{100-text-20-audio-20-video}
    \label{fig:config-c}
\end{subfigure}

\vspace{0.5em}

\begin{subfigure}[b]{0.31\textwidth}
    \centering
    \begin{tikzpicture}
    \begin{axis}[
        xmin=0, xmax=110,
        ymin=0.68, ymax=0.695,
        yticklabel style={/pgf/number format/fixed, /pgf/number format/fixed zerofill, /pgf/number format/precision=3, font=\scriptsize},
        yticklabel style={font=\tiny},
        xtick={20, 40, 60, 80, 100},
        xticklabels={$k$=20,$k$=40,$k$=60,$k$=80,$k$=100},
        xticklabel style={rotate=50, anchor=east, font=\tiny},
        width=\linewidth,
        height=0.75\linewidth,
        grid style=dashed,
        ymajorgrids=true,
        xmajorgrids=true,
        mark size=1.2pt,
        line width=0.8pt,
        tick label style={font=\small},
        label style={font=\small}
    ]
    \addplot[color=orange, mark=pentagon] coordinates {
        (10,0.689)(20,0.689)(30,0.69)(40,0.687)(50,0.69)
        (60,0.691)(70,0.687)(80,0.685)(90,0.685)(100,0.691)
    };
    \end{axis}
    \end{tikzpicture}
    \caption*{20-text-100-audio-100-video}
    \label{fig:config-d}
\end{subfigure}
\hfill
\begin{subfigure}[b]{0.31\textwidth}
    \centering
    \begin{tikzpicture}
    \begin{axis}[
        xmin=0, xmax=110,
        ymin=0.585, ymax=0.61,
        yticklabel style={/pgf/number format/fixed, /pgf/number format/fixed zerofill, /pgf/number format/precision=3, font=\scriptsize},
        yticklabel style={font=\tiny},
        xtick={20, 40, 60, 80, 100},
        xticklabels={$k$=20,$k$=40,$k$=60,$k$=80,$k$=100},
        xticklabel style={rotate=50, anchor=east, font=\tiny},
        width=\linewidth,
        height=0.75\linewidth,
        grid style=dashed,
        ymajorgrids=true,
        xmajorgrids=true,
        mark size=1.2pt,
        line width=0.8pt,
        tick label style={font=\small},
        label style={font=\small}
    ]
    \addplot[color=purple, mark=star] coordinates {
        (10,0.59)(20,0.605)(30,0.604)(40,0.599)(50,0.605)
        (60,0.605)(70,0.596)(80,0.594)(90,0.596)(100,0.597)
    };
    \end{axis}
    \end{tikzpicture}
    \caption*{20-text-100-audio-20-video}
    \label{fig:config-e}
\end{subfigure}
\hfill
\begin{subfigure}[b]{0.31\textwidth}
    \centering
    \begin{tikzpicture}
    \begin{axis}[
        xmin=0, xmax=110,
        ymin=0.68, ymax=0.70,
        yticklabel style={/pgf/number format/fixed, /pgf/number format/fixed zerofill, /pgf/number format/precision=3, font=\scriptsize},
        yticklabel style={font=\tiny},
        xtick={20, 40, 60, 80, 100},
        xticklabels={$k$=20,$k$=40,$k$=60,$k$=80,$k$=100},
        xticklabel style={rotate=50, anchor=east, font=\tiny},
        width=\linewidth,
        height=0.75\linewidth,
        grid style=dashed,
        ymajorgrids=true,
        xmajorgrids=true,
        mark size=1.2pt,
        line width=0.8pt,
        tick label style={font=\small},
        label style={font=\small}
    ]
    \addplot[color=brown, mark=otimes] coordinates {
        (10,0.692)(20,0.688)(30,0.686)(40,0.685)(50,0.686)
        (60,0.687)(70,0.688)(80,0.688)(90,0.687)(100,0.695)
    };
    \end{axis}
    \end{tikzpicture}
    \caption*{20-text-20-audio-100-video}
    \label{fig:config-f}
\end{subfigure}

\vspace{0.5em}

\begin{subfigure}[b]{0.31\textwidth}
    \centering
    \begin{tikzpicture}
    \begin{axis}[
        xmin=0, xmax=110,
        ymin=0.705, ymax=0.730,
        yticklabel style={/pgf/number format/fixed, /pgf/number format/fixed zerofill, /pgf/number format/precision=3, font=\scriptsize},
        yticklabel style={font=\tiny},
        xtick={20, 40, 60, 80, 100},
        xticklabels={$k$=20,$k$=40,$k$=60,$k$=80,$k$=100},
        xticklabel style={rotate=50, anchor=east, font=\tiny},
        width=\linewidth,
        height=0.75\linewidth,
        grid style=dashed,
        ymajorgrids=true,
        xmajorgrids=true,
        mark size=1.2pt,
        line width=0.8pt,
        tick label style={font=\small},
        label style={font=\small}
    ]
    \addplot[color=cyan!60!black, mark=oplus] coordinates {
        (10,0.717)(20,0.711)(30,0.715)(40,0.724)(50,0.713)
        (60,0.711)(70,0.719)(80,0.722)(90,0.718)(100,0.721)
    };
    \end{axis}
    \end{tikzpicture}
    \caption*{complex-20-20-20-10-10-10-10}
    \label{fig:config-g}
\end{subfigure}

\end{minipage}%
}
\caption{F-Micro scores across different number of clusters ($k$) for the missing configurations.}
\label{fig:effects-of-number-of-clusters}
\end{figure*}

A natural question arising from the UL4M4 framework is how sensitive the imputation quality — and consequently the downstream task performance — is to the choice of the number of clusters $k$. A method that requires careful tuning of $k$ for each dataset or missing-modality configuration would be considerably less practical than one that is robust across a broad range of values. To investigate this, we evaluate UL4M4 on CMU-MOSI across ten values of $k$ spanning a wide range, under all missing-modality configurations, and report F-Micro scores in Figure~\ref{fig:effects-of-number-of-clusters}. In addition, Table~\ref{tab:ul4m4-varying-k} complements the figure by reporting both MSE and F-Micro for three representative values of $k$ side by side, providing a more precise quantitative picture that the compressed y-axes of the plots — necessary to reveal within-configuration variation — can obscure.

The overarching finding from both the figure and the table is that UL4M4 is remarkably stable across the entire range of $k$ evaluated. Within each configuration, F-Micro scores fluctuate only within a very narrow band across all tested values, and no systematic trend of improvement or degradation with increasing $k$ is apparent. The same stability is reflected in the MSE columns of Table~\ref{tab:ul4m4-varying-k}: across the three representative values of $k$, the differences in error are negligible for virtually every configuration. This consistency holds across both the easier settings — where the majority of modalities are available — and the harder ones where text availability is severely reduced, which are otherwise the most challenging configurations for the model.

This robustness has an important practical implication: practitioners do not need to invest significant effort in tuning $k$ on a held-out validation set, and a moderate default value selected without dataset-specific tuning is sufficient to obtain performance representative of the method's full capability. It also suggests that the cluster-guided imputation mechanism does not rely on the cluster structure being highly refined — even a relatively coarse partitioning of the embedding space captures enough of the shared multimodal structure to produce effective imputations.
\begin{table}[t]
\centering
\caption{UL4M4 results with varying number of clusters ($k$) on CMU-MOSI.}
\label{tab:ul4m4-varying-k}
\small
\begin{tabular}{l@{\hspace{14pt}}ccc@{\hspace{14pt}}ccc}
\toprule
 & \multicolumn{3}{c}{MSE $\downarrow$} & \multicolumn{3}{c}{F-Micro $\uparrow$} \\
\cmidrule(lr){2-4}\cmidrule(lr){5-7}
\scriptsize{Configuration}
  & \rotatebox{90}{UL4M4/$k$=40}
  & \rotatebox{90}{UL4M4/$k$=70}
  & \rotatebox{90}{UL4M4/$k$=100}
  & \rotatebox{90}{UL4M4/$k$=40}
  & \rotatebox{90}{UL4M4/$k$=70}
  & \rotatebox{90}{UL4M4/$k$=100} \\
\midrule
100-text-100-audio-100-video & 1.01  & 1.01  & 1.01  & 0.81 & 0.81 & 0.81 \\
100-text-100-audio-20-video  & 1.04 & 1.05 & 1.04 & 0.80 & 0.81 & 0.81 \\
100-text-20-audio-100-video  & 0.99 & 0.98 & 0.99 & 0.81 & 0.82 & 0.81 \\
100-text-20-audio-20-video   & 1.05 & 1.02 & 1.05 & 0.81 & 0.81 & 0.81 \\
20-text-100-audio-100-video  & 1.88 & 1.88 & 1.89 & 0.69 & 0.69 & 0.69 \\
20-text-100-audio-20-video   & 2.19 & 2.20 & 2.19 & 0.60 & 0.60 & 0.60 \\
20-text-20-audio-100-video   & 1.91 & 1.90 & 1.89 & 0.69 & 0.69 & 0.70 \\
complex-20-20-20-10-10-10-10 & 1.61 & 1.61 & 1.59 & 0.72 & 0.72 & 0.72 \\
\bottomrule
\end{tabular}
\end{table}


\subsection{Implementation Details}
\label{section:implementation-details}
    For CMU-MOSI, UL4M4 uses BERT encoder with weights 'bert-base-uncased' for text, WavLM-Large encoder with 'microsoft/wavlm-large' weights for audio, and CLIP vision encoder with 'openai/clip-vit-base-patch32' for video frames, as its frozen encoders. As the problem is originally regression problem with ranging from -3 (strongly negative) to +3, for UL4M4, the task head $g: \mathbb{R}^{d_f} \to \mathbb{R}$ is a linear projection followed by a bounded activation. Specifically, the prediction for sample $i$ is computed as:
    \begin{equation} \label{eq:task_head}
    \hat{y}_i = \text{HardTanh}_{[-3,\,3]}\!\left( W^{\text{head}} \mathbf{F}_{\mathcal{S},i} + b^{\text{head}} \right),
    \end{equation}
    where $W^{\text{head}} \in \mathbb{R}^{1 \times d_f}$ and $b^{\text{head}} \in \mathbb{R}$ are learnable parameters, and $\text{HardTanh}_{[-3,3]}$ clamps the output to the interval $[-3, 3]$, defined as:
    \begin{equation} \label{eq:hardtanh}
    \text{HardTanh}_{[-3,\,3]}(x) =
    \begin{cases}
    -3 & \text{if } x < -3, \\
    x  & \text{if } -3 \leq x \leq 3, \\
    3  & \text{if } x > 3.
    \end{cases}
    \end{equation}
    For regression, the training objective is the mean squared error (MSE) loss:
    \begin{equation} \label{eq:mse}
    \mathcal{L}_{\text{task}} = \frac{1}{N} \sum_{i=1}^{N} \left( y_i - \hat{y}_i \right)^2,
    \end{equation}
    where $y_i \in \mathbb{R}$ is the ground-truth label and $\hat{y}_i$ is the predicted scalar for sample $i$, with $\mathcal{L}_{\text{task}}$ driving end-to-end optimization of all trainable components.
    
    For MM-IMDb, UL4M4 uses frozen CLIP (ViT-B/32) encoders for both image and text modalities, producing embeddings of equal dimensionality for each. The  downstream task is multi-label genre classification, so the task head consists 
    of a multi-layer feedforward network with dropout and ReLU activations, trained with binary cross-entropy loss. Model selection during training is based on the best validation F1-Micro score across epochs.

    For Sleep-EDF, no suitable pretrained encoder exists for raw EEG signals, so UL4M4 incorporates an additional Stage~0 in which a lightweight 1-D convolutional autoencoder is trained from scratch on the raw signal windows in a fully unsupervised manner. Once pretraining is complete, the decoder is discarded and the encoder is frozen; the resulting embeddings are then passed through the standard UL4M4 Stages~1 and~2. The task head for the five-class sleep-stage classification is a small feedforward network trained with 
    weighted cross-entropy loss, with model selection based on best validation accuracy.
\section{Conclusion}
\label{section:Conclusion}

In this paper, we introduced UL4M4, a novel framework for addressing the missing-modality problem in multimodal learning. By leveraging unsupervised clustering with modality-specific normalization and a partial-modality distance metric, UL4M4 enables robust imputation of missing feature embeddings in a task-independent manner. This decoupling allows seamless integration with any downstream supervised prediction architecture, supporting arbitrary numbers of modalities and highly flexible missing patterns at the sample level.

Our extensive experiments across three diverse benchmarks — multimodal sentiment analysis (CMU-MOSI), multi-label movie genre classification (MM-IMDb), and polysomnographic sleep staging (Sleep-EDF) — under diverse and severe missing regimes demonstrate UL4M4's effectiveness and generality. In configurations where large fractions of modality slots are absent, UL4M4 achieves competitive or superior performance compared to both general-purpose baselines and task-specific methods such as MissModal, which is purpose-built for sentiment analysis. Notably, UL4M4 is the first method to report F1-Micro scores consistently above 0.7 on CMU-MOSI under such severe missing-modality conditions. The method also proves largely insensitive to the choice of cluster number $k$ across a wide range of values, relieving practitioners of the burden of careful hyperparameter tuning.

UL4M4 advances multimodal learning by providing a general, lightweight solution that handles real-world data incompleteness without compromising downstream task accuracy. Future directions include extending the framework to streaming or online settings where cluster centres must be updated incrementally, exploring learned or adaptive distance metrics to further improve imputation quality, and investigating fully unsupervised variants that operate without any downstream task supervision.

\bibliographystyle{unsrt}
\bibliography{references}

\end{document}